\documentclass[10pt,twocolumn]{article}

\usepackage{times}
\usepackage{epsfig}
\usepackage{graphicx}
\usepackage{amsmath}
\usepackage{amssymb}
\usepackage{siunitx}
\usepackage{subfig}
\usepackage{multirow}
\usepackage{dblfloatfix}
\usepackage{booktabs}
\usepackage[cm]{fullpage}

\usepackage[pagebackref=true,breaklinks=true,colorlinks,bookmarks=false,citecolor=black]{hyperref}

\begin{document}

\title{Fast Sparse ConvNets}
\date{}

\setlength{\tabcolsep}{2.5em}
\renewcommand\arraystretch{1.2}

\author{
\large
\begin{tabular}{c c c c}
\textbf{Erich Elsen}\thanks{These authors contributed equally to this work.} &
\textbf{Marat Dukhan}\footnotemark[1] &
\textbf{Trevor Gale} \footnotemark[1]~~\thanks{Work done as part of the Google AI Residency} &
\textbf{Karen Simonyan} \\
DeepMind & Google & Google & DeepMind\\
\multicolumn{4}{c}{\small\{eriche, maratek, tgale, simonyan\}@google.com}\\
\end{tabular}
}

\setlength{\tabcolsep}{1.0em}

\maketitle

\begin{abstract}
Historically, the pursuit of efficient inference has been one of the driving forces behind research into new deep learning architectures and building blocks. Some recent examples include: the squeeze-and-excitation module~\cite{squeezeexcite}, depthwise separable convolutions in Xception \cite{Xception}, and the inverted bottleneck in MobileNet v2~\cite{mobilenetv2}. Notably, in all of these cases, the resulting building blocks enabled not only higher efficiency, but also higher accuracy, and found wide adoption in the field. In this work, we further expand the arsenal of efficient building blocks for neural network architectures; but instead of combining standard primitives (such as convolution), we advocate for the replacement of these dense primitives with their sparse counterparts. While the idea of using sparsity to decrease the parameter count is not new~\cite{Thimm95evaluatingpruning}, the conventional wisdom is that this reduction in theoretical FLOPs does not translate into real-world efficiency gains. We aim to correct this misconception by introducing a family of efficient sparse kernels for ARM and WebAssembly, which we open-source for the benefit of the community as part of the XNNPACK~\cite{xnnpack} library. Equipped with our efficient implementation of sparse primitives, we show that sparse versions of MobileNet v1, MobileNet v2 and EfficientNet architectures substantially outperform strong dense baselines on the efficiency-accuracy curve. On Snapdragon 835 our sparse networks outperform their dense equivalents by $1.3-2.4\times$ -- equivalent to approximately one entire generation of MobileNet-family improvement. We hope that our findings will facilitate wider adoption of sparsity as a tool for creating efficient and accurate deep learning architectures.
\end{abstract}

\section{Introduction}

\begin{figure*}
\begin{center}
    \begin{subfloat}[][Top-1 Accuracy vs. FLOPs]{
      \centering
      \includegraphics[width=0.45\linewidth]{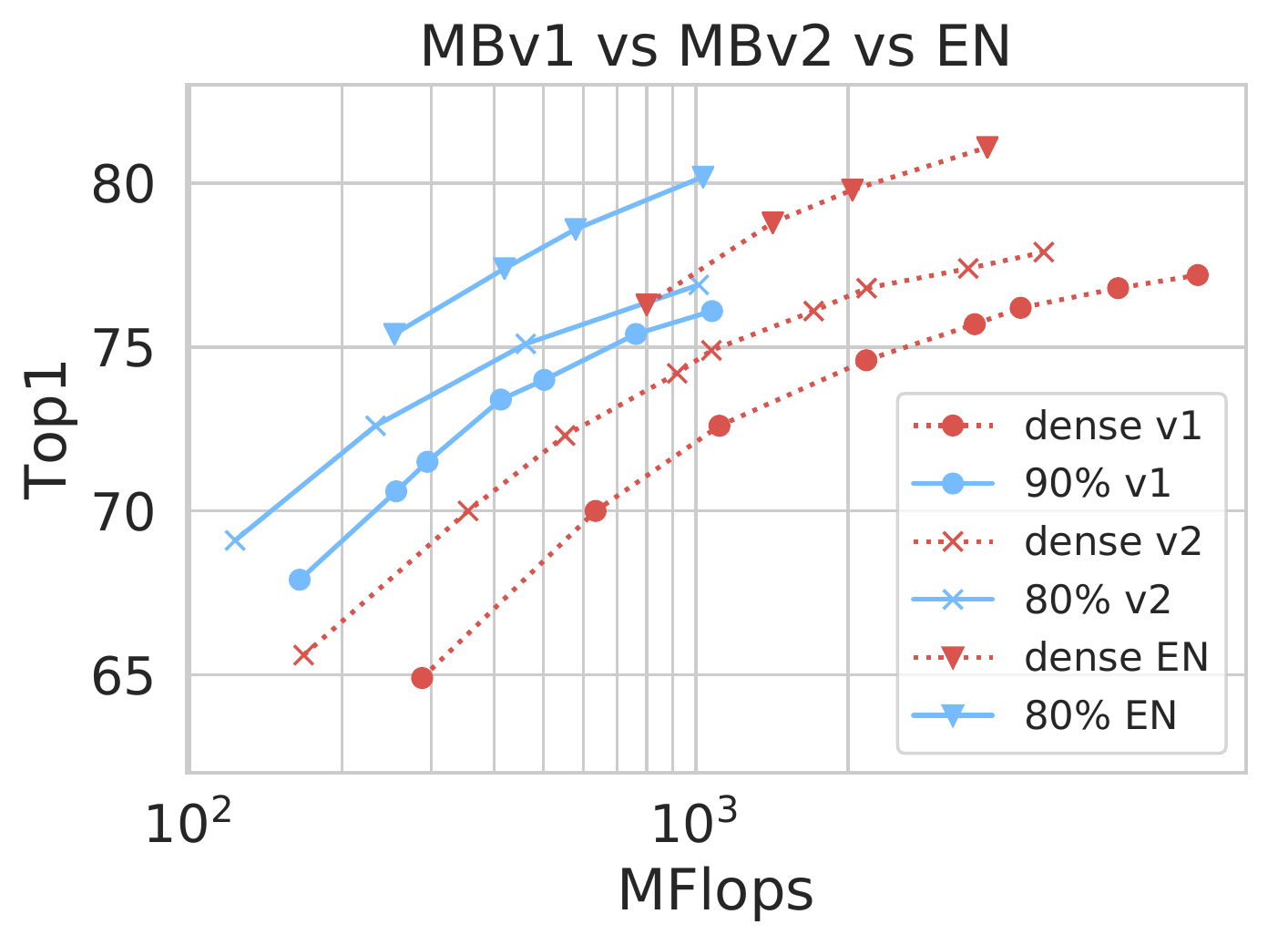}}
    \end{subfloat}
    ~
    \begin{subfloat}[][Top-1 Accuracy vs. Parameter Count]{
      \centering
      \includegraphics[width=0.45\linewidth]{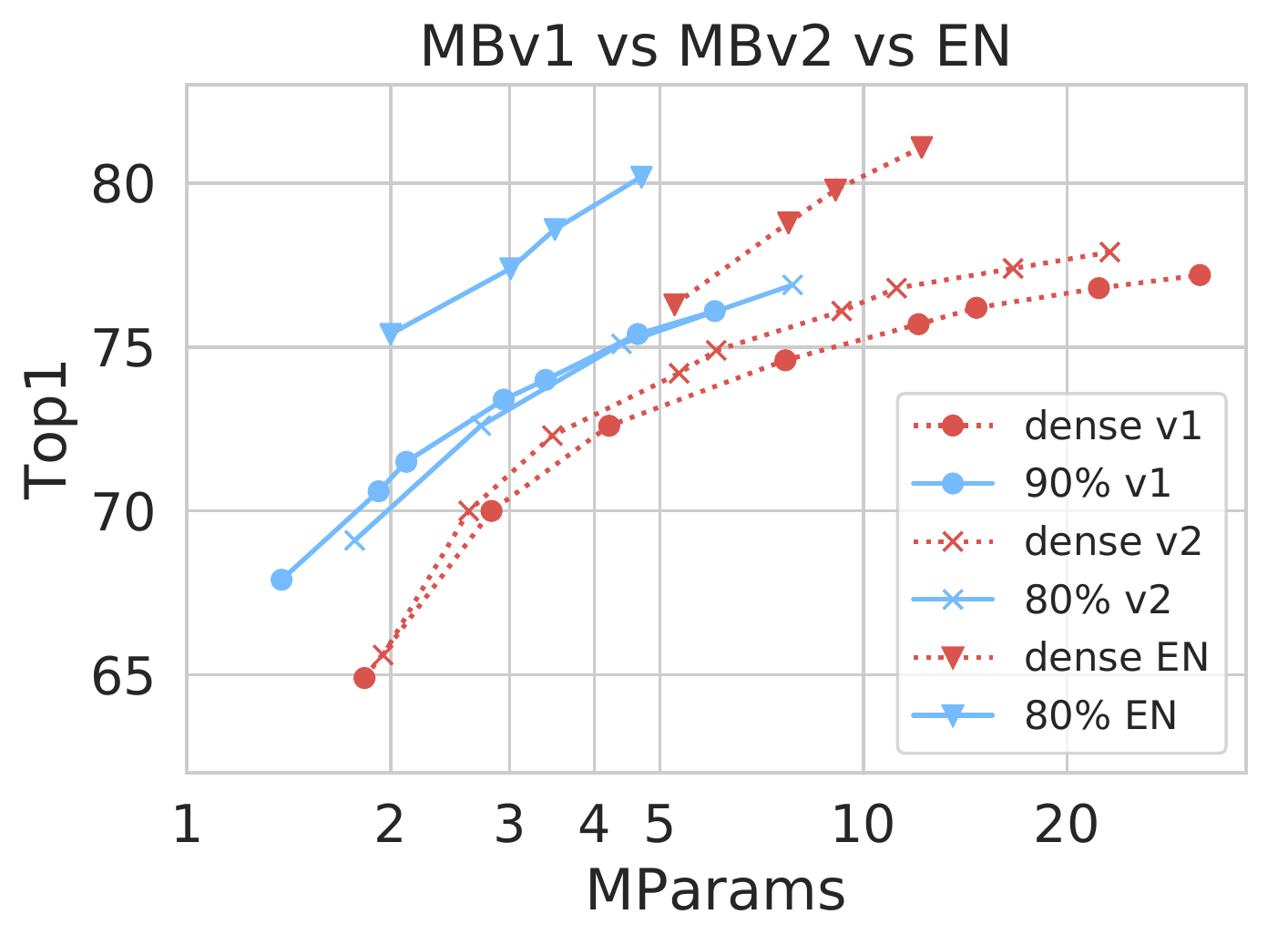}}
    \end{subfloat}
\end{center}
    \caption{MobileNet v1 and v2 and EfficientNet models.  Sparse models: blue (solid), dense models: red (dotted). Sparse models include the cost of storing the location of non-zeros for sparse tensors as a bitmask converted back into parameter count.  That is every 32 values in the bitmask contributes one ``parameter''.}
    \label{fig:FLOP_curve}
\end{figure*}

Convolutional neural networks (CNNs) have proven to be excellent at solving
a diverse range of tasks~\cite{bhandare2016applications}. Standard network architectures are used in classification,
segmentation, object detection and generation tasks~\cite{GANProgress, long2015fully, zhao2019object}.  Given their wide utility, there has been significant effort to design
efficient architectures that are capable of being run on mobile and other
low power devices while still achieving high classification accuracy on
benchmarks such as ImageNet~\cite{imagenet}.  
For example, MobileNets~\cite{mobilenetv1, mobilenetv2} employ the depthwise separable convolutions introduced by Sifre in~\cite{Sifre14ecolepolytechnique} to significantly reduce resource requirements over 
previous architectures.  Inference time, FLOPs and parameter counts in these architectures
are dominated by the 1$\times1$ convolutions, which directly map to matrix-matrix multiplications.

Weight sparsity is generally known to lead~\cite{cheng2017survey} to theoretically smaller and more computationally efficient (in terms of number of floating-point operations) 
models, but it is often disregarded
as a practical means of accelerating models because of the misconception that sparse operations cannot be fast enough to achieve actual speedups during inference. To address this misconception and firmly establish sparsity as a tool in the deep learning practitioner's arsenal, we introduce fast kernels for 
sparse matrix-dense matrix multiplication (SpMM) specifically targeted at the acceleration of sparse neural networks. The main distinction of our SpMM kernel from prior art~\cite{SpGEMM_KNL, YangOwens_GPU_SpMM} is that we focus on a different point in the design space.  While prior work focused on extremely sparse problems (typically \textgreater 99\%, found in scientific and graph problems), we target the sparsity range of 70-95\%, more common when inducing weight sparsity in neural networks.
As a result our kernels significantly outperform the kernels generated by the TACO compiler~\cite{taco_compiler} and the Intel MKL~\cite{intel-mkl}.

Using these kernels, we demonstrate the effectiveness of weight sparsity across three generations of 
MobileNet~\cite{mobilenetv1, mobilenetv2, efficientnet} architectures. Sparsity leads to an improvement of approximately one whole generation in each architecture, with sparse EfficientNets being significantly more efficient than all previous models. These models represent a new generation of efficient CNNs, which reduces inference times by $1.3-2.4\times$, parameter counts by over $2\times$ and number of floating-point operations (FLOPs) by up to $3\times$ relative to the previous generations.

\section{Related Work}

Improvements in convolutional network architectures~\cite{alexnet, vggnet, rn50, densenet}, as measured by increased 
classification accuracy on benchmark tasks such as ImageNet~\cite{imagenet}, 
have generally been concomitant with increases in model parameter counts, FLOPs 
and memory requirements.  Recently this evolution has led to networks found through neural architecture
search~\cite{NASNet, AmoebaNet} which can achieve over 82\% top-1 accuracy, but require
nearly 25 GFLOPs for one inference.

Given these prohibitive inference costs, there have been many lines of work attempting
to improve CNN efficiency, which is often defined as one of three metrics:
\begin{enumerate}
    \item Inference speedup on real hardware
    \item Theoretical speedup through FLOPs reduction
    \item Model size reduction
\end{enumerate}

\begin{figure*}
\begin{center}
    \includegraphics[width=\textwidth]{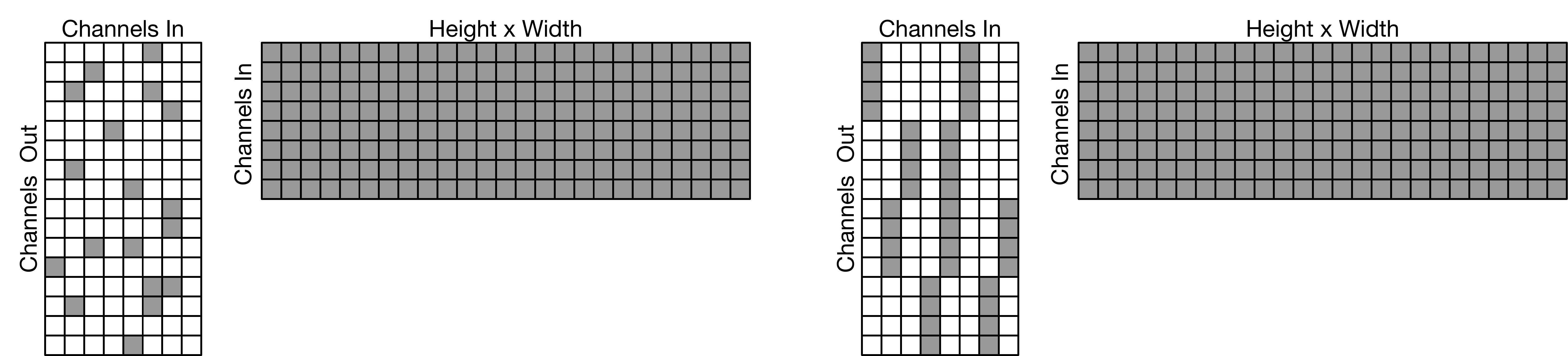}
\end{center}
    \caption{Sparse 1x1 Convolution as SpMM. Left: Unstructured sparsity (or block size 1). ~Right: Output channel block size of 4}
    \label{fig:SpMM_1x1}
\end{figure*}

These axes are neither parallel nor orthogonal.  The effect of (3) and (2) on (1) in 
particular can be quite complicated and highly varied depending on the hardware in question.

The MobileNet family of architectures~\cite{mobilenetv1, mobilenetv2} 
has focused on improving efficiency by taking advantage of the depthwise separable convolutions introduced
in~\cite{Sifre14ecolepolytechnique}, which can be thought of as a hand-crafted sparsification
of full convolutions with a predefined sparse topology, and which are responsible for the parameter efficiency of these
architectures.  MobileNet v1 (MBv1) used layers of $1\times1$ convolutions followed by depthwise convolutions.  MobileNet v2 (MBv2) introduced the inverted residual block which consists of a $1\times1$ convolution expanding the channel count, a depthwise convolution on the expanded channel count, and then a $1\times1$ convolution reducing the parameter count. Across MobileNet architectures, the depthwise convolutions account for only a small fraction of the total FLOPs, parameters, and inference time of these models.  In MBv1, they account for less than 2\% of the total FLOPs and in MBv2 less than 3\%.

\begin{figure*}[!b]
\begin{center}
    \includegraphics[width=\textwidth]{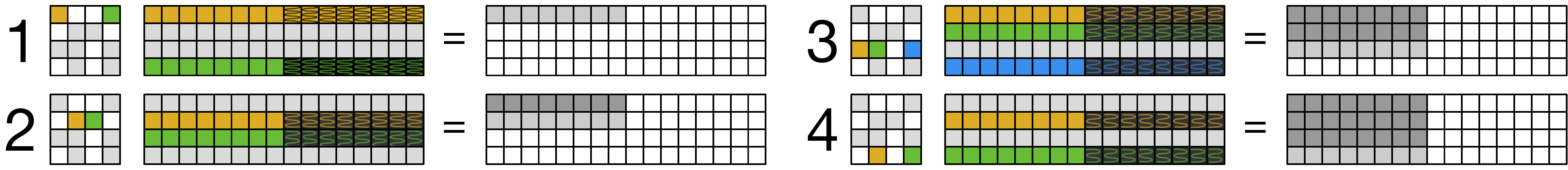}
\end{center}
    \caption{Visualization of the memory reads and writes of our algorithm. In step 1, we load 8 spatial locations simultaneously for each of the non-zero weights in the first row of the weight matrix. We also prefetch the values that will be needed for the next set of columns (shown in same color but hatched).  We multiply each scalar weight by its corresponding row, accumulate the results, and in the end write them out. Step 2 performs the same calculation for the next output channel. After steps 1 and 2, all values for these spatial locations are in the cache, so future loads in steps 3 and 4 will be fast, despite being random access.}
    \label{fig:kernel_scheme}
\end{figure*}

A different line of work attempted to make more efficient CNNs by directly 
pruning the weights of full convolutional filters accompanied by the necessary
inference kernels~\cite{SparseCNN_Intel_Park16, Liu2015SparseCNN}.  \cite{SparseCNN_Intel_Park16} was not able to accelerate $1\times1$ convolutions, \cite{Liu2015SparseCNN} did not attempt it. The latter also required generating a new set of kernels for each instance of a model, which is often impractical for deployment. Due to the difficultly of accelerating sparse computation, channel pruning approaches have been preferred~\cite{morphnet, variational-information-bottleneck, thinet, l0-regularization, fisher-pruning, automatic-model-compression}. These approaches prune away entire filters leaving the final model dense, and function more as an architecture search over channel counts.

Full Neural Architecture Search has also been applied directly to architectures resembling MBv2 resulting in MobileNet v3~\cite{mobilenetv3},
FBNet~\cite{FBNet}, and EfficientNet~\cite{efficientnet}.

Alternatively, factorizations of the 1$\times$1 convolutions have been considered
in ShuffleNet~\cite{shufflenet} and Learnable Butterfly Factorizations~\cite{butterfly_stanford}.  
ShuffleNet factorizes the weight matrix into a product of a permutation matrix
and block diagonal matrix. Butterfly Factorizations factorize the weight matrix into
a sequence of permutation matrices and weight matrices with special structure that
can represent many common $O(Nlog N)$ transforms such as Fast Fourier Transforms.

Work in Text-to-Speech (TTS)~\cite{wavernn} demonstrated that increasing sparsity and concomitant
increase in state size in RNN models lead to increased model quality for a given non-zero parameter count.  They additionally demonstrated fast block-sparse matrix-vector (SpMV) multiplication routines necessary for RNN inference.

\section{Methods}

To understand how to design the most efficient convolutional models, we investigate both how to construct and train sparse MBv1, MBv2 and EfficientNet models and also the performance of our SpMM kernels.

\subsection{Sparsifying Networks}

We train on the ImageNet~\cite{imagenet} dataset with standard augmentation and report top-1 accuracies on the provided 50k example validation set. To make the networks sparse we use the gradual magnitude pruning technique of~\cite{to-prune-or-not}.

We do not prune the first full convolution at the beginning of all three networks. Its overall contribution to the parameter count, FLOP count, and runtime is small and does not warrant introducing a new sparse operation.  Instead, we implement a dense convolutional kernel which takes as input the image in the standard HWC layout and outputs the CHW layout consumed by the sparse operations in the rest of the network. In HWC layout, the values for different channels corresponding to one spatial location are adjacent in memory. In CHW layout, the values of all the spatial locations for one channel are adjacent in memory.

We also do not prune the squeeze-excitation~\cite{squeezeexcite} blocks in EfficientNet as they contribute \textless 1\% of the total FLOPs to the dense model. The last fully-connected layer in all models also contributes insignificantly (\textless 1\%) to the total FLOP count, but does contribute a significant fraction (20-50\%) of total parameters, especially after the rest of the model is
pruned. As we are concerned with maximizing top-1 accuracy for a given runtime, we do not prune the final layer in MobileNet v1 and v2 as doing so leads to a small decrease in top-1 accuracy.  Standard EfficientNets do not scale the number of filters in the last convolution by the width of the model, however we find that when introducing sparsity it is beneficial to do this; in all sparse EfficientNet models we double the units from 1280 to 2560. We also find that it is possible to make the fully-connected layer sparse without loss of accuracy in EfficientNet, so we do so.

\begin{figure*}
\begin{center}
    \begin{subfloat}[][MB v1 ARM NEON]{
      \centering
      \includegraphics[width=.45\linewidth]{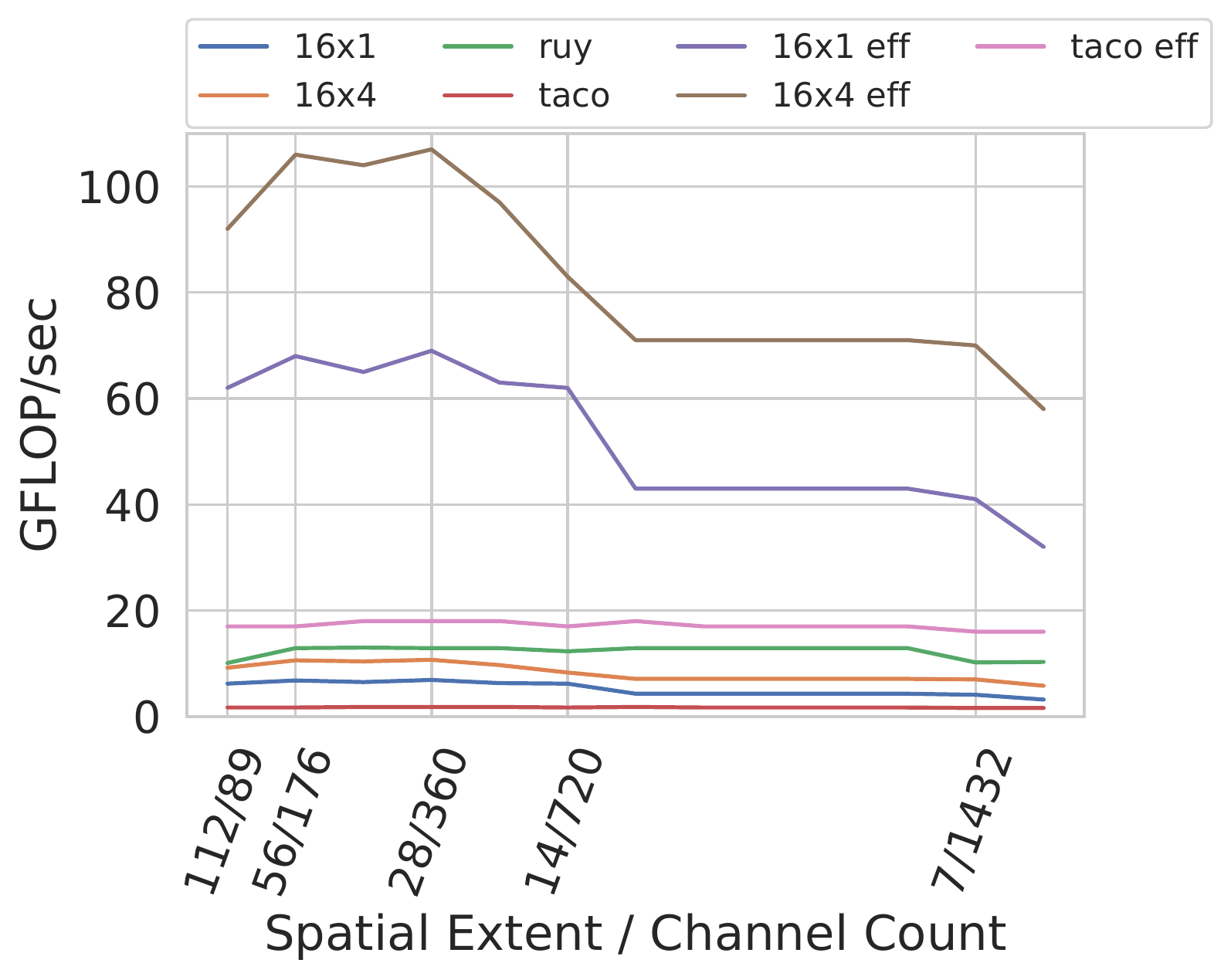}}
    \end{subfloat}
    ~
    \begin{subfloat}[][MB v2 ARM NEON]{
      \centering
      \includegraphics[width=.45\linewidth]{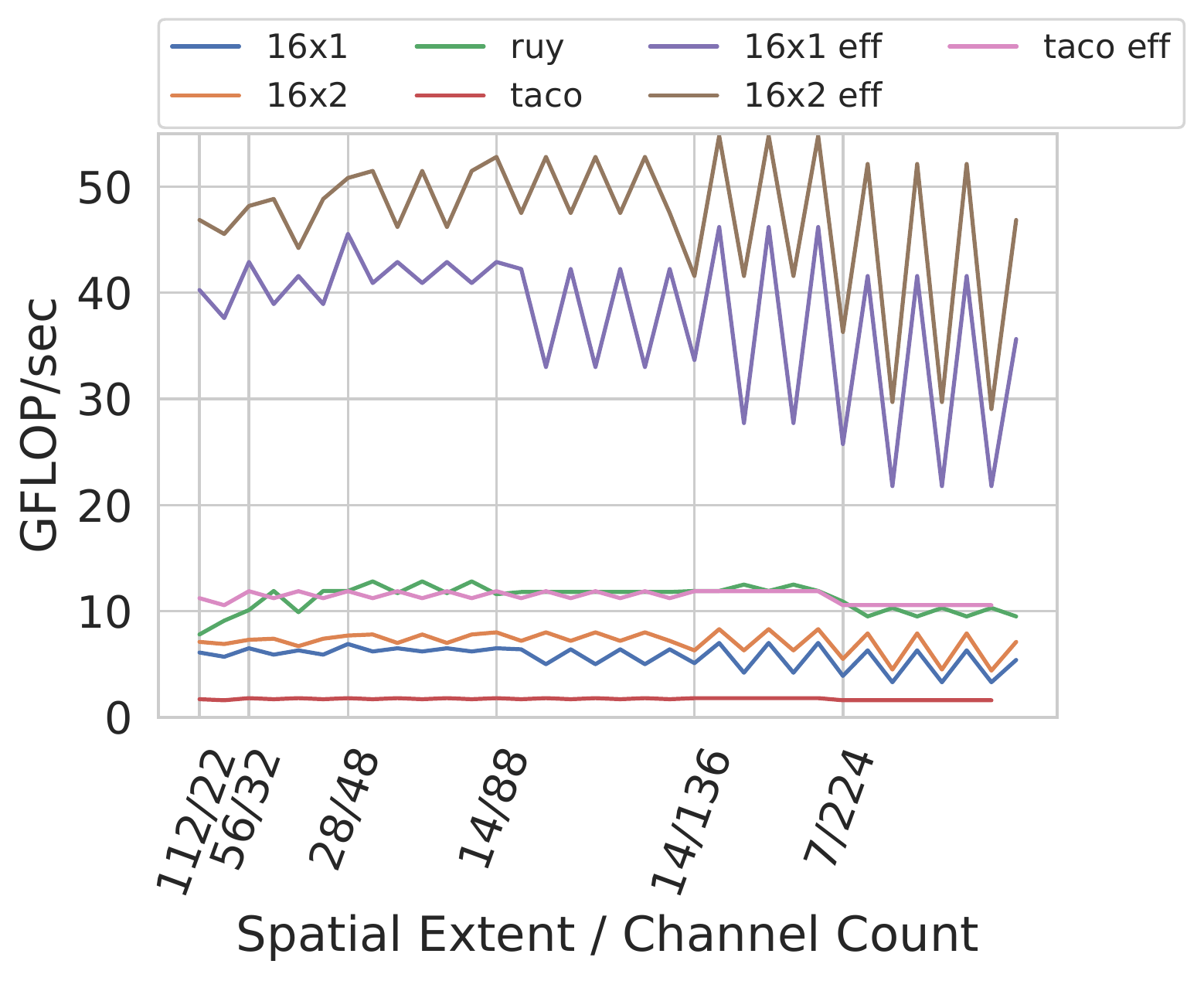}}
    \end{subfloat}
\end{center}
    \caption{FLOPs with increasing layer depth. All measurements taken on a Snapdragon (SD) 835. Effective assumes 90\% sparse MBv1 and 85\% sparse MBv2 models.}
    \label{fig:flop_with_depth}
\end{figure*}

\subsection{Kernel Implementation} A diagram of the $1\times1$ convolution as a SpMM is seen in figure~\ref{fig:SpMM_1x1}.  Our scheme requires activation tensors be stored in CHW format, in contrast to dense mobile inference libraries~\cite{GEMMLOWP, QNNPACK, RUY} which favor HWC.

There are three key insights enabling the high performance of our kernels: 
\begin{enumerate}
    \item While the weight matrix is sparse, the activation matrix is dense. This means that we can perform vector loads from the activation matrix and process multiple spatial locations simultaneously.
    \item By processing the matrix in the right order we can keep values that will be randomly accessed in the L1 cache, from which random access is fast and constant time.
    \item When the number of input channels is small enough, prefetching from the activations can further reduce cache misses. 
\end{enumerate}  Figure~\ref{fig:kernel_scheme} shows the memory read and write patterns of a few steps of the kernel. The figure shows 8 elements being processed together for visualization but 16 is more natural for the actual implementation as it corresponds to one cache line.  The outer loop is over columns and the inner loop is over rows; this allows each strip of 16 spatial locations in the activations to remain in the L1 cache until it is no longer needed. In figure~\ref{fig:kernel_scheme} steps 1 and 2 prime the cache, while subsequent steps 3 and 4 load all right hand side values from the L1 cache.

\begin{table*}[h!]
\begin{center}
    \begin{tabular}{l|c|ccccccc}
    \toprule
         &  Model & Width & Top-1 & Mega  & Mega   & Time  & Time  & Time \\
         &        &       &       & FLOPs & Params & SD835 & SD670 & Wasm\\
         \midrule
        \multirow{2}{*}{MBv1}
        & Dense & 1.0       & 70.9 & 1120 & 4.30 & 125 & 106 & 271\\
        & Sparse & 1.4      & \textbf{72.0} & \textbf{268} & \textbf{2.28} & \textbf{58} & \textbf{64} & \textbf{97} \\\midrule
        \multirow{2}{*}{MBv1}
        & Dense & .75       & 68.4 & 636 & 2.59 & 73 & 64 & 170 \\
        & Sparse & 1.0      & 68.4 & \textbf{146} & \textbf{1.48} & \textbf{31} & \textbf{34} & \textbf{56} \\ \midrule
        \multirow{2}{*}{MBv1}
        & Dense & .5        & 63.3 & 290 & 1.34 & 36 & 33 & 96 \\
        & Sparse & .75      & \textbf{64.4} & \textbf{90} & \textbf{1.30} & \textbf{21} & \textbf{21} & \textbf{36} \\
        \midrule
        \multirow{3}{*}{MBv2}
        & Dense & 1.4       & \textbf{75.0} & 1110 & 6.06 & 150 & 129 & 319 \\
        & Sparse & 2.0      & 74.5 & \textbf{406} & 4.24 & \textbf{93} & \textbf{91} & \textbf{155} \\
        & Sparse* & 1.8      & \textbf{74.9} & 411 & \textbf{4.13} & 102 & 108 & \textbf{155} \\
        \midrule
        \multirow{2}{*}{MBv2}
        & Dense & 1.0       & 71.8 & 580 & 3.47 & 83 & 74 & 197 \\
        & Sparse & 1.4      & 72.0 & \textbf{220} & \textbf{2.68} & \textbf{54} & \textbf{53} & \textbf{95}\\
        \midrule
        \multirow{3}{*}{MBv2}
        & Dense & .75       & 69.8 & 375 & 2.61 & 64 & 57 & 154 \\
        & Sparse & 1.15     & \textbf{70.2} & 165 & 2.11 &  40 & 39 & \textbf{74} \\ 
        & CA Sparse$\ddagger$  & 1.0    & 69.7 & \textbf{119} & \textbf{1.73} & \textbf{33} & \textbf{35} & - \\ \midrule
        \multirow{2}{*}{MBv2}
        & Dense & .5        & 65.4 & 182 & 2.05 & 33 & 30 & 92 \\
        & Sparse & .80      & 65.2 & \textbf{90} & \textbf{1.66} & \textbf{26} & \textbf{24} & \textbf{41} \\ \midrule
        \multirow{2}{*}{EN}
        & Dense  &  EN-b0      & 76.8 & 730 & 5.28 & 158 & 148 & - \\
        & Sparse &  EN-b1      & 76.7 & \textbf{220} & \textbf{3.07} & \textbf{110} & \textbf{118} & - \\
        \bottomrule
    \end{tabular}
\end{center}
    \caption{Comparison of dense and sparse model sizes, flops, and inference speeds. All input image sizes are 224x224.  Sparse MBv1 models are 90\% sparse in every layer, Sparse MBv2 models are 85\% sparse. In sparse MBv1 models, layer 12 uses a block size of 4. This is almost as efficient as the models in~\ref{sec:model_design} and matches the top-1 scores of the dense models more closely.  In sparse MBv2 models, layers 11 and onward use a block size of 2.  The finally fully connected layer in all models is dense. All times are in milliseconds. Dense times on ARM are measured using TensorFlow Lite.  Web Assembly results are measured on an Intel W-2135, dense times use Intel's webml-polyfill~\cite{webml-polyfill} library$^\dagger$.  Sparse parameter counts include the overhead of sparsity storage as a bitmask for each sparse layer. EN-b1 is 85\% sparse and unstructured, final FC layer is sparse.\\
    *This model is 80\% sparse in all layers (except final fully-connected) and uses a block size of 1 everywhere.\\
    $\ddagger$~This is the cache aware MBv2 architecture described in section~\ref{sec:cacheaware}. It uses a block size of 1 throughout.\\
    $^\dagger$We also tried WebDNN~\cite{hidaka2017webdnn}, but it does not support some operations necessary to run MobileNet and EfficientNet models.}
    \label{tab:time_results}
\end{table*}

In addition to the vectorization in the $HW$ dimension, taking advantage of small amounts of structure in the weight matrix can offer significant performance boosts by increasing data reuse after values are loaded into registers. Constraining the sparsity pattern so that multiple output or input channels all share the same zero/non-zero pattern creates `blocks' in the weight matrix (see figure~\ref{fig:kernel_scheme} right).  Blocks in the output channel dimension  
allow for more data reuse than blocks in the input channel dimension.  Experiments (see figure~\ref{fig:block_effect}) show that either choice has the same effect on accuracy, so we implement output channel blocking with sizes of 2 and 4. Our nomenclature for kernels is to give their spatial vectorization width followed by the output channel block size -- \verb 16x2  ~means 16 pixels and 2 output channels are processed in the inner loop.

We implement the ARM kernels in C with NEON intrinsics unlike current production libraries~\cite{GEMMLOWP, QNNPACK, RUY} which rely on expert-optimized assembly. All the SpMM kernels used in this work are available as part of XNNPACK~\cite{xnnpack}. 

\subsection{Library} We will provide a library that can run sparse models trained with the model pruning library in TensorFlow~\cite{tensorflow2015-whitepaper}.  This includes conversion from a dense representation to a Block Compressed Sparse Row (BCSR)-like representation suitable for inference. In addition to the high performance $1\times1$ convolutions, we also provide all supporting CHW kernels -- depthwise convolutions, global average pooling and a $3\times3$ stride-2 dense convolution that takes the standard HWC format as input and outputs CHW -- necessary for running all three generations of models.  While we provide high performance versions of these kernels, we do not detail them here; they are also available as part of XNNPACK~\cite{xnnpack}. Times for these kernels are included in end-to-end measurements.

\subsection{WebAssembly}

In some settings it is useful to run inference (and even training) of deep neural networks directly in web browsers.  This is supported now by multiple frameworks including WebDNN~\cite{hidaka2017webdnn}, Tensorflow.js~\cite{tensorflowjs} and Webml-polyfill~\cite{webml-polyfill}.  Frameworks generally support using WebAssembly (Wasm)~\cite{webassembly} to run on CPUs or WebGL to run on GPUs.  In this work we target web assembly backends and show that sparse vision networks significantly outperform their dense counterparts in this setting.  In this setting our kernel decomposition strategy is similar, however due to current lack of vectorization support in Wasm, they are converted to scalar instructions.  Due to the smaller register file, we find that unrolling by 8 instead of 16 is optimal.  Scalar versions of all kernels are also provided as part of XNNPACK~\cite{xnnpack}.

\section{Results}

In the main text we mainly include plots for MBv1 and MBv2 due to space limitations. EfficientNets generally follow the same trends as MBv2 models, plots for EfficientNet can be found in the supplementary material.  Table~\ref{tab:time_results} contains the main results comparing inference times for models achieving approximately the same top-1.

To breakdown the end-to-end results, first we reveal performance results for our SpMM kernels, then we show how the networks respond to sparsity and then finally how we combined this information to find the models with the lowest inference time.

\subsection{ARM Kernel Performance}
We use Ruy~\cite{RUY}, the current TensorFlow Lite ARM64 backend written largely in hand-coded assembly, as the dense baseline. For a sparse baseline we use the kernel generated by the TACO compiler~\cite{taco_compiler}. We present results by plotting the FLOPs achieved at each layer in the model, with increasing depth to the right in figure~\ref{fig:flop_with_depth}. For MBv1 we use a width multiplier of 1.4 and 90\% sparse and for MBV2 we use a width multiplier of 1.4 and 85\% sparse as these configurations approximately match the top-1 accuracy of the width 1 dense models. The kernel variants that process 16 spatial locations at a time (e.g. \verb 16x1 , etc.) are the highest performing and all reported numbers are from these kernel variants. TACO results should be compared with the \verb 16x1 ~kernels.

The raw performance of the sparse kernels falls in the range of 40--90\% of the dense kernels.  And as they must do much less work, when taking the sparsity of the layer into account, the effective FLOPs are in the 2--7$\times$ range. In MBv1 performance falls significantly in the last two layers of the model when the number of channels (1024) causes the size of one ``strip'' of spatial locations to exceed the size of the L1 cache.  In MBv2 the sawtooth pattern is caused by the alternating expand and contract operations.  The performance is higher for the expand kernels due to greater data reuse of each ``strip'' that is brought into the L1 cache.  The performance of the contract operations drop significantly once the number of expanded channels in each residual block exceeds the size of the L1 cache; a smaller drop is observed when it exceeds the size of half of the L1 cache as this prevents effective prefetching.

\begin{figure*}
\begin{center}
    \begin{subfloat}[][MBv1 90\% Sparse]{
      \centering
      \includegraphics[width=0.45\linewidth]{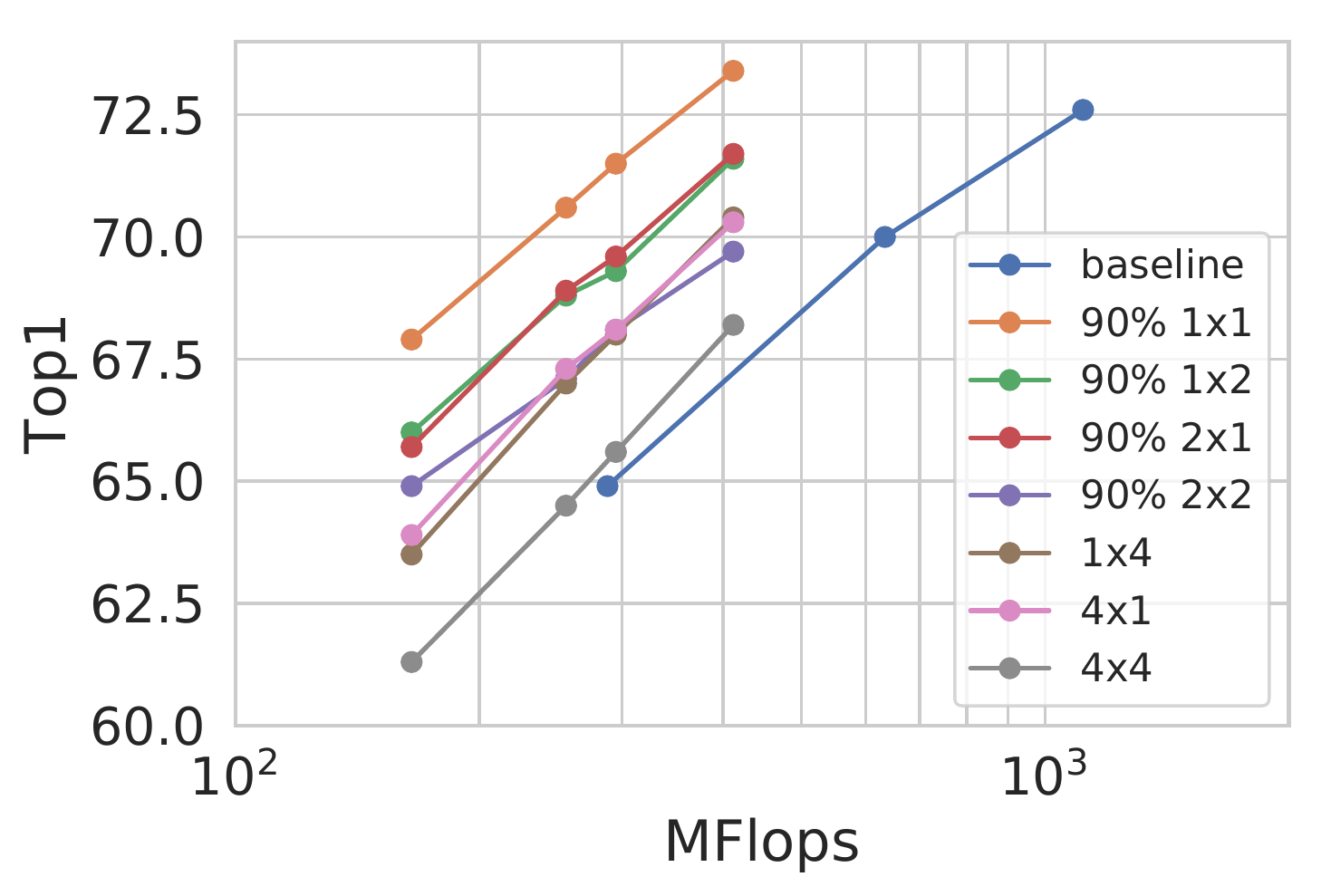}}
    \end{subfloat}
    ~
    \begin{subfloat}[][MBv2 80\% Sparse]{
      \centering
      \includegraphics[width=0.45\linewidth]{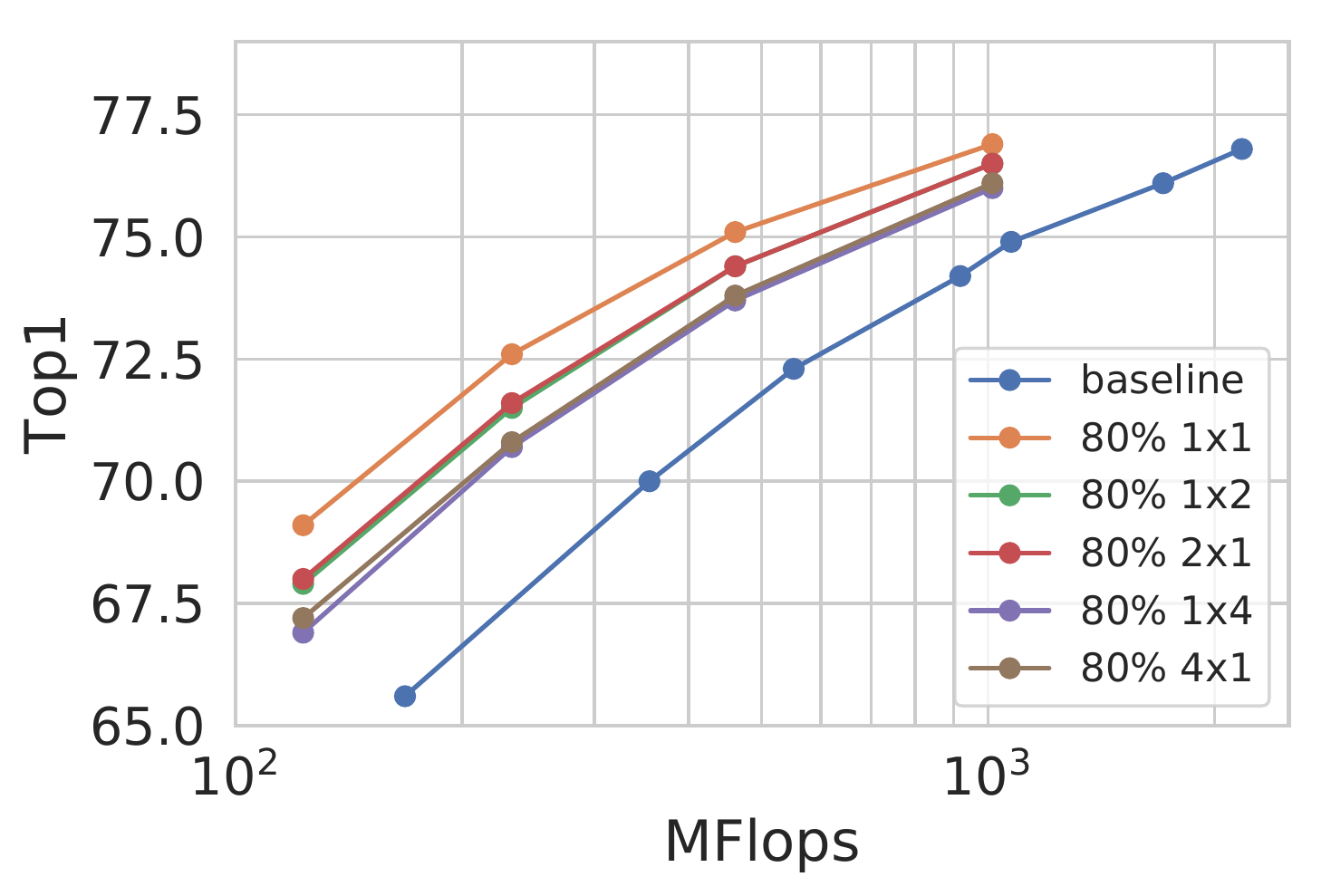}}
    \end{subfloat}
\end{center}
    \caption{Effect of block size on top-1 accuracy. It only matters how many elements are in a block, the configuration is unimportant.}
    \label{fig:block_effect}
\end{figure*}

\subsection{Model Performance}\label{sec:experiments}

The hyper-parameters used to train MBv1 and MBv2 are listed in table~\ref{tab:hparams}, they were found with a grid search on dense models with a width multiplier of 1.0 to reproduce the original results, which used RMSProp~\cite{rmsprop}, with SGD with momentum. The same hyper-parameters are used to train sparse models. This change allows us to match or exceed the reported accuracies with only 38,000 iterations of training. 

\begin{table}[h!]
    \centering
    \begin{tabular}{l|r|r}
    \toprule
         & MBv1 & MBv2 \\ \midrule
        learning rate  & $.35 * 16 = 5.6$ & $.24 * 16 = 3.84$ \\
        momentum       & 0.9 & 0.92  \\
        l2 coefficient & 5e-5 & 4e-5  \\ \bottomrule
    \end{tabular}
    \caption{Hyper-parameters for MBv1 and MBv2 training.  Learning rates are specified in a reduced space and then multiplied by a factor of 16 due to the batch size (4096). The learning rate schedule is a linear ramp for the first 8 epochs to the maximum value followed by step-wise decay at epochs 40, 75 and 95 by a factor of ten.
    }
    \label{tab:hparams}
\end{table}

The hyper-parameters used to train EfficientNet are largely unmodified from their code release, with the exception of extending training from 350 to 650 epochs and increasing the learning rate decay exponent to .985 from .97 so that the learning rate decays more slowly. These changes do not improve the dense baseline.

We induce sparsity in MBv1 and MBv2 by extending training by a factor (along with learning rate anchor points) of four, we find this increases the performance of the sparse models, but not the baselines.  We start the sparsification process at iteration $7,000*4=28,000$ and stop at $28,000*4=112,000$ with a pruning frequency of 2,000. For EfficientNet we start at iteration 23,000 and end at iteration 105,000, also with a pruning frequency of 2,000.  See \cite{to-prune-or-not} for the meaning of these hyper-parameters.

We train on the ImageNet~\cite{imagenet} dataset with standard data augmentation. Top-1 accuracies are reported on the validation set with center single-crops.

To understand the effect of block size, we plot in figure~\ref{fig:block_effect} accuracy against flops for different block sizes.  In these plots, every sparse tensor in the network uses the same output channel block size. The tradeoff for block sparsity only appears to involve how many elements are in each block, and not their configuration. For example, in MBv1, the $1\times4$, $4\times1$ and $2\times2$ curves all lie on top of one another. The loss in accuracy due to blocking seems to decrease slightly for larger width models.

\begin{figure*}
\begin{center}
    \begin{subfloat}[][MBv1]{
      \centering
      \includegraphics[width=0.45\linewidth]{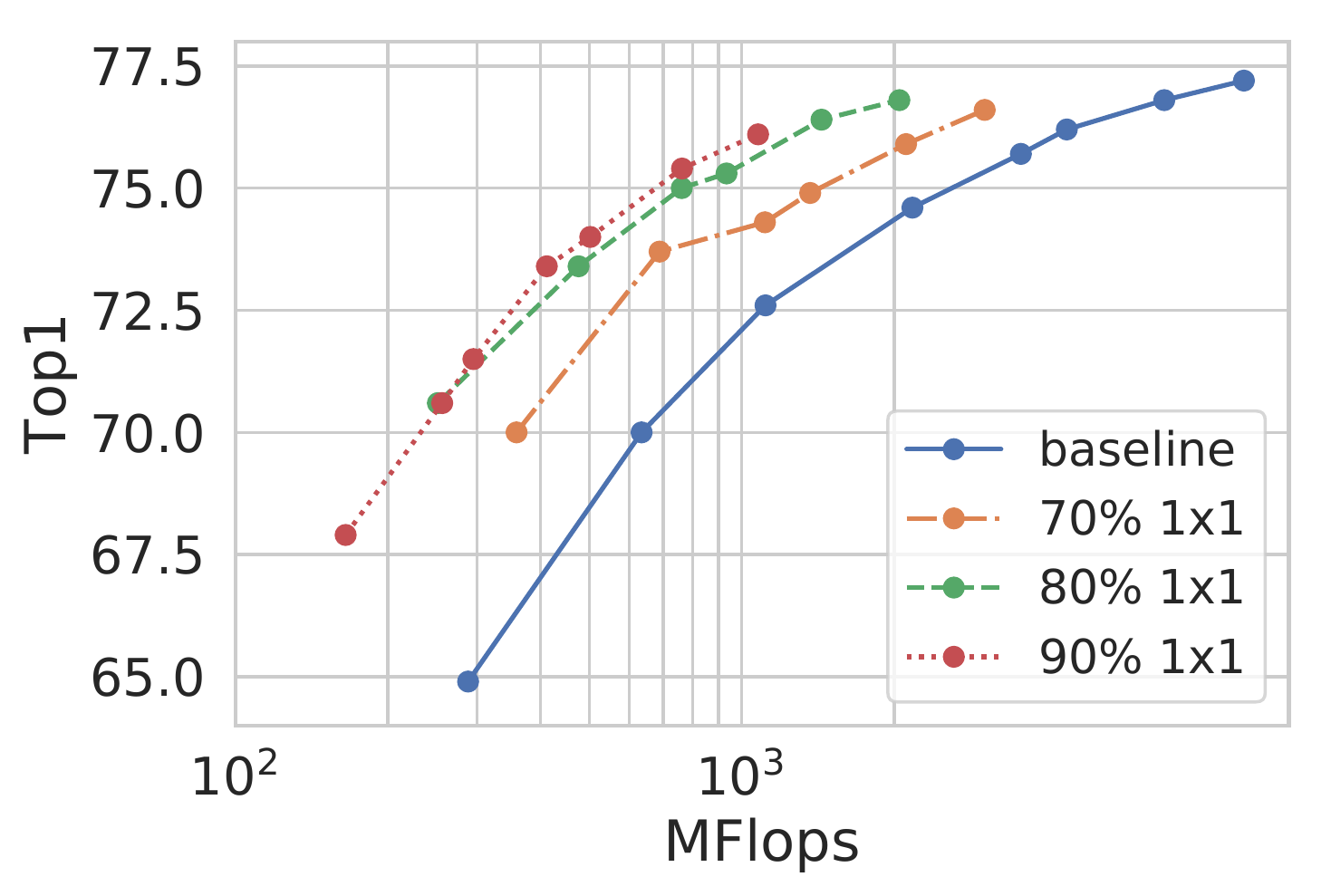}}
    \end{subfloat}
    ~
    \begin{subfloat}[][MBv2]{
      \centering
      \includegraphics[width=0.45\linewidth]{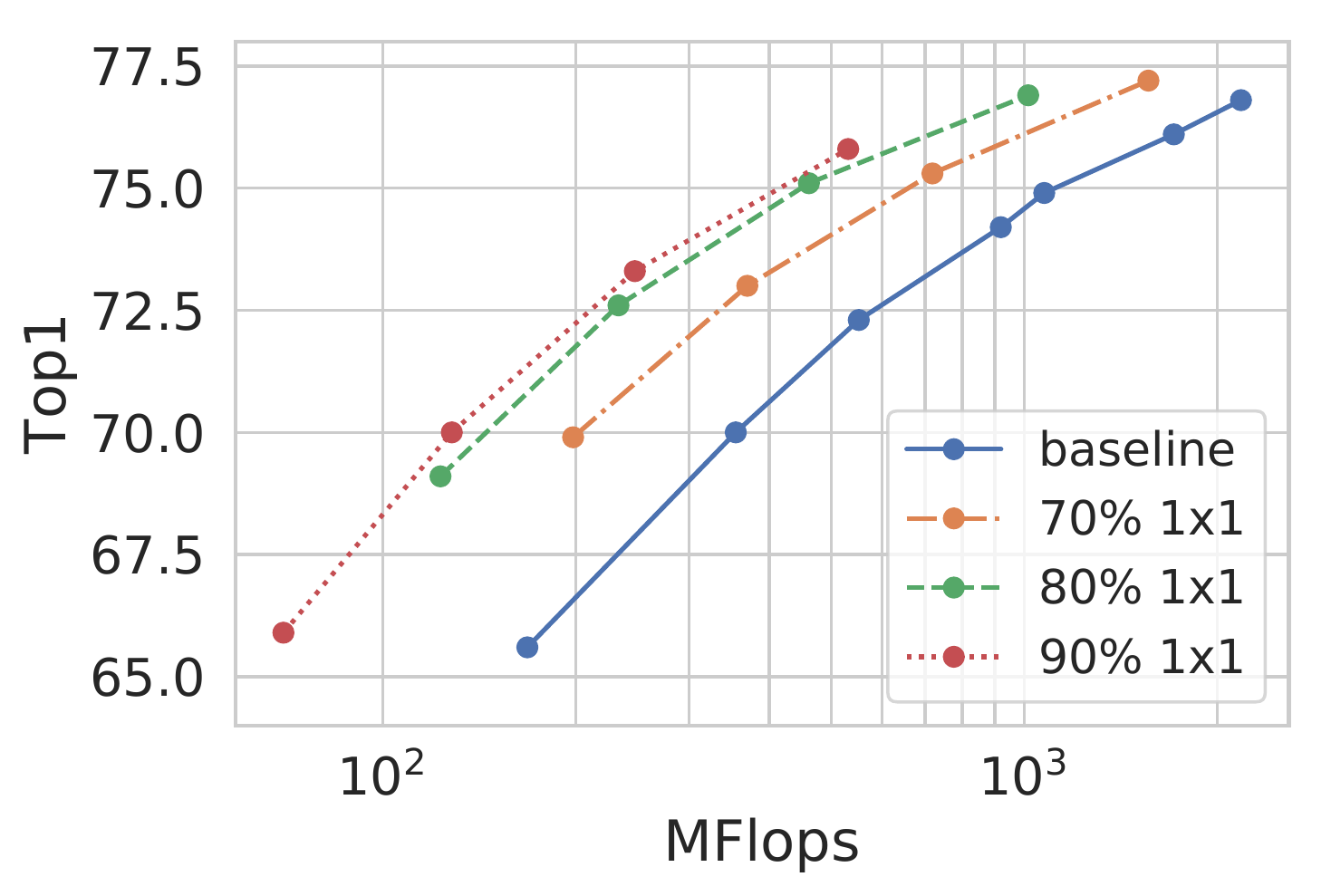}}
    \end{subfloat}
\end{center}
    \caption{Effect of sparsity on top-1 accuracy.  The sparser a model is, the fewer flops it requires to achieve a given Top-1 accuracy.}
    \label{fig:sparsity_effect}
\end{figure*}

To understand how the sparsity level affects the efficiency of the models, we train models at 70\%, 80\% and 90\% unstructured sparsity which is constant throughout the model. The results are plotted in figure~\ref{fig:sparsity_effect}. MBv1 and MBv2 are more efficient the more sparse they become, confirming that the results of~\cite{wavernn} hold not just for RNNs, but also for convolutional models as well.

In figure~\ref{fig:FLOP_curve} we plot Top-1 accuracy vs. FLOPs for all three generations of sparse and dense models. MobileNet v1 is 90\% sparse, the other models are 80\% sparse. A sparse MBv1 exceeds MBv2 in terms of FLOP and parameter efficiency; a sparse MBv2 matches EfficientNet in terms of FLOP and parameter efficiency; and a sparse EfficientNet exceeds all other models in both categories.

\begin{figure*}
\begin{center}
    \begin{subfloat}[][MBv1]{
      \centering
      \includegraphics[width=0.45\linewidth]{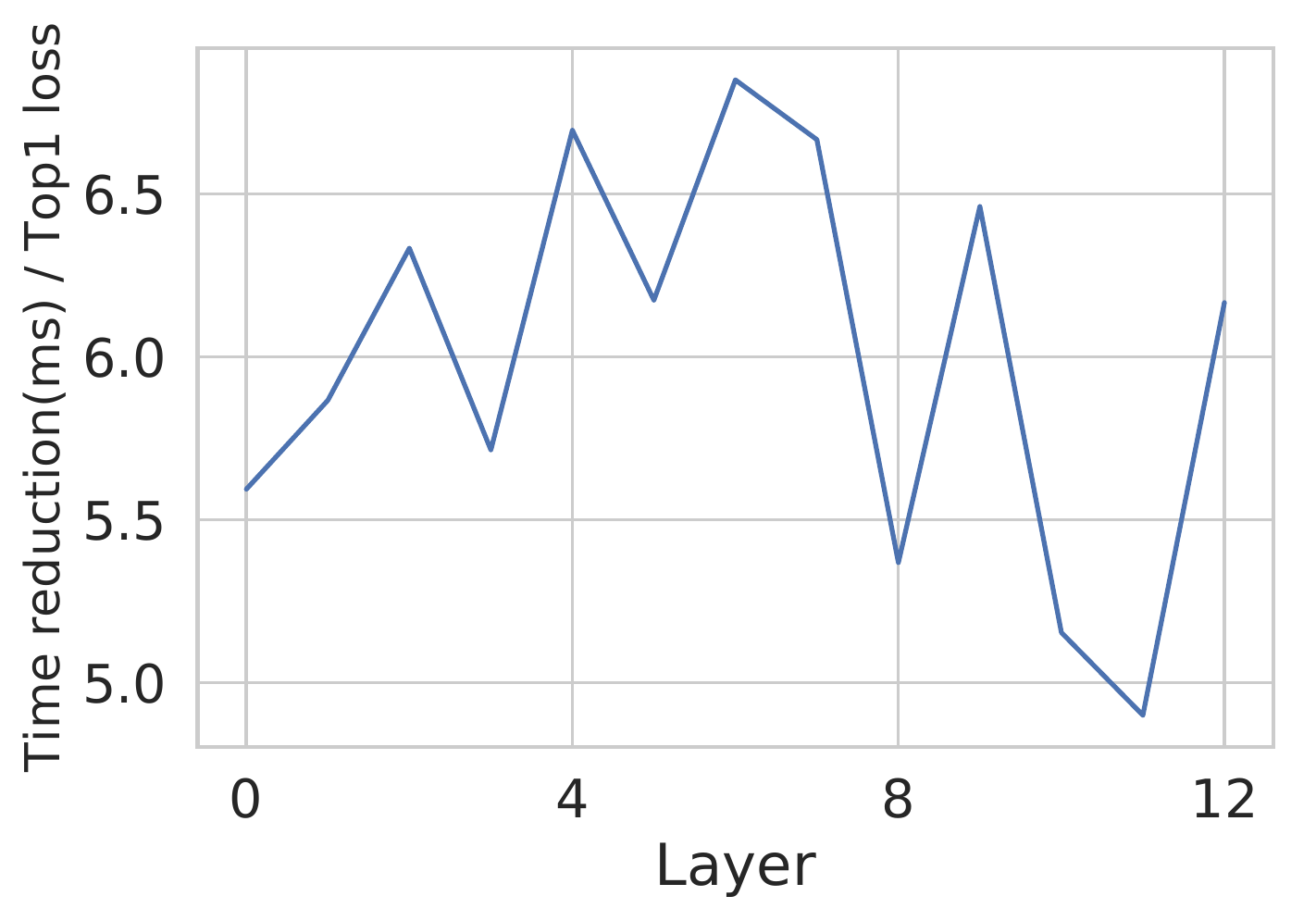}}
    \end{subfloat}
    ~
    \begin{subfloat}[][MBv2]{
      \centering
      \includegraphics[width=0.45\linewidth]{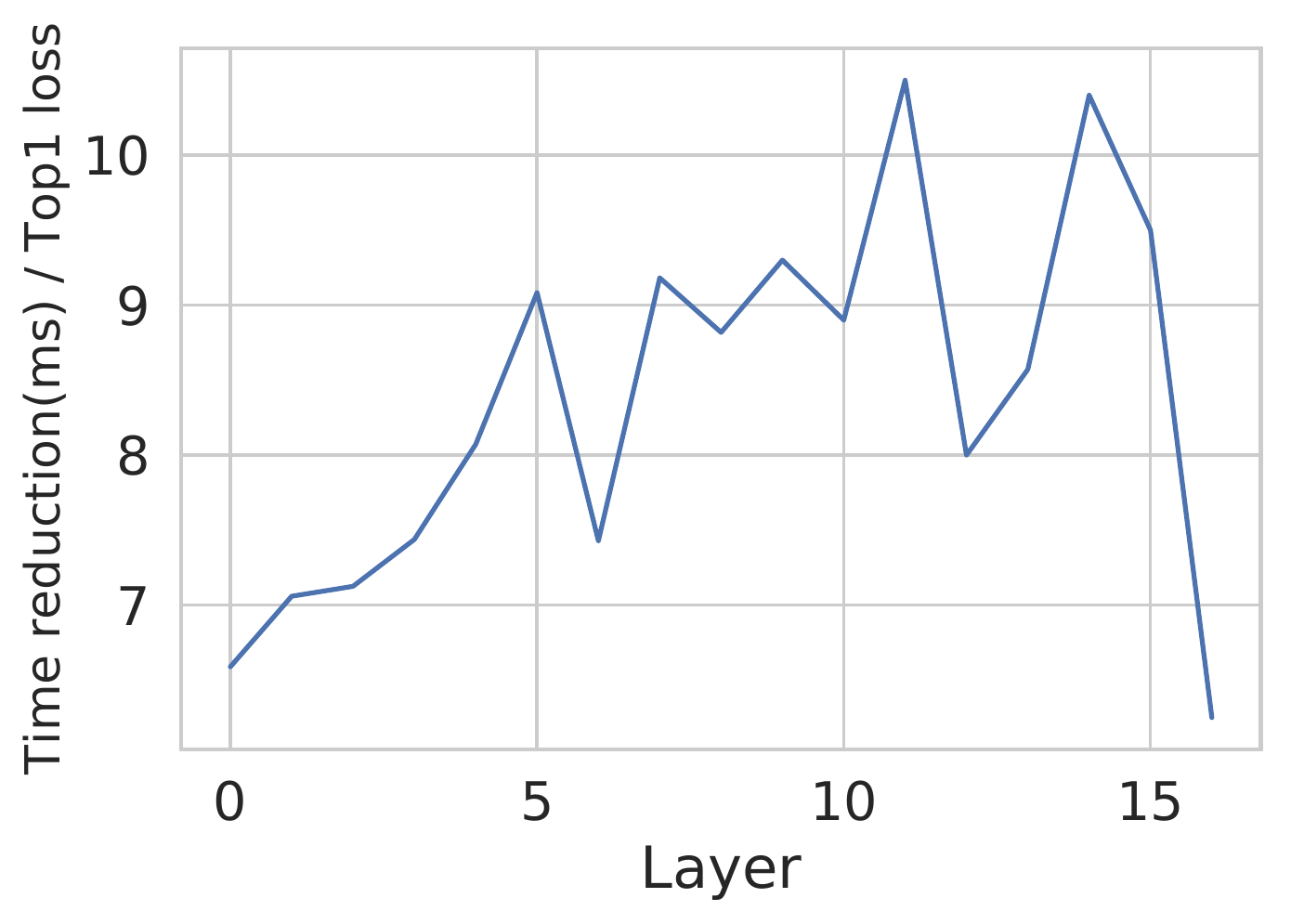}}
    \end{subfloat}
\end{center}
    \caption{Efficiency of models with layer N and onward blocked. The x-axis corresponds to turning that layer and all following layers to block size 4, the prior layers are unstructured.  The y-axis is the efficiency of making this change over an unstructured model given as a ratio where the numerator is the speedup of changing the block(s) from unstructured to block size 4 and the denominator is the decrease in top-1 accuracy that occurs by making this change.}
    \label{fig:model_selection}
\end{figure*}

\subsection{Model Design for Block Size}
\label{sec:model_design}
To design the models with the best top-1 accuracy vs. inference time frontiers we make the following assumptions to reduce the search space:
\begin{enumerate}
    \item We leave the models themselves unchanged.
    \item We induce the same level of sparsity in all $1\times1$ convolutions.
\end{enumerate}

Then we do a search at width multiplier 1.4 over $N$ models when there are $N$ residual blocks in a model. An x-axis location of $n$ corresponds to a model in which the first $n$ residual blocks are unstructured and the last $N-n$ residual blocks have an output channel block size of 4. We train each model, note its top-1 accuracy and then measure its inference time. From this we can calculate the ratio of inference time reduction relative to a fully unstructured model and top-1 lost, which are plotted in figure~\ref{fig:model_selection}. We choose the model with the highest ratio and train models at all widths with this choice. This amounts to making layers 6 and deeper blocked in MBv1 models and layers 11 and deeper blocked in MBv2.

\subsection{Cache Aware Model Design}
\label{sec:cacheaware}

The sawtooth pattern in figure~\ref{fig:flop_with_depth} is due to the large number of channels during the contract phase causing the size of stripe to exceed the size of the L1 cache.  One possible solution, common with dense kernels, would be to split the matrix into N pieces such that each piece only accesses a number of channels that fit within the cache.  However, this introduces additional complexity in the software -- in the sparse case it would require repacking the matrix, which isn't required in the dense case.  As another possible solution we examine a simple modification to the MBv2 architecture to make it cache aware.

The inverted residual block introduced by MBv2 expands the number of channels before the depthwise convolution and then reduces them afterwards.  The expansion factor is fixed at 6 for all layers.  Once there are more than 512 channels after expansion, the size of the 32Kb L1 cache is exceeded during contraction.  Additionally, once there are more than 256 channels after expansion, the data exceeds half of the cache precluding effective use of pre-fetching during contraction.  To design an architecture that is aware of these constraints we take MBv2 and increase the expansion factor of early layers while reducing the expansion factor as the number of channels increases so that the number of expanded channels still fits within approximately half of the cache size.  To compensate for the decrease in capacity this would otherwise cause, we also increase the depth of the model by adding one more layer with 32 channels, two more with 64 channels and three more each with 96 and 160 channels.  The final architecture is in table~\ref{tab:cache_aware_mbv2}.

\begin{table}[h]
    \centering
    \begin{tabular}{llcccc}
        \toprule
        Input &  Operation & e & c & n & s\\ \midrule
        $224^2 \times 3$ & conv2d      & - & 16   & 1 & 2 \\
        $112^2 \times 16$& Bottleneck  & 1 & 16   & 1 & 1 \\
        $112^2 \times 16$& Bottleneck  & 8 & 24   & 2 & 2 \\
        $56^2 \times 24$&  Bottleneck  & 8 & 32   & 4 & 2 \\
        $28^2 \times 32$&  Bottleneck  & 4 & 64   & 6 & 2 \\
        $14^2 \times 64$&  Bottleneck  & 3 & 96   & 6 & 1 \\
        $7^2 \times  96$&  Bottleneck  & 2 & 160  & 6 & 2 \\
        $7^2 \times  160$& Bottleneck  & 2 & 320  & 1 & 1 \\
        $7^2 \times  320$& conv2d 1x1  & - & 1280 & 1 & 1 \\
        $7^2 \times  1280$& gavgpool   & - & - & 1 & - \\
        $1^2 \times  1280$& conv2d 1x1 & - & k & 1 & - \\ \bottomrule
    \end{tabular}
    \caption{Architecture of cache aware MBv2 optimized for our sparse kernels. Each block is repeated n times and contains c channels that are expanded by a factor of e.  The initial convolution in each block has stride s, all others have stride 1.}
    \label{tab:cache_aware_mbv2}
\end{table}

\subsection{Wallclock Times}

Table~\ref{tab:time_results} contains the timings for running our sparse models on a single big core of two different processors, a Snapdragon 835 and a Snapdragon 670. We compare them with MBv1 and MBv2 models from their official repositories~\cite{TFmbv1, TFmbv2} run on the dense-inference TF Lite framework with the standard Ruy backend.\footnote{The dense EfficientNet b0 model we trained ourselves using the official code with a slight modification to make exporting a TF-lite model easier.}  Model files for all sparse models in table~\ref{tab:time_results} are available \href{https://github.com/google-research/google-research/tree/master/fastconvnets}{here}.

Surprisingly, in the presence of sparsity MBv1 is approximately as efficient as MBV2 suggesting that a full Neural Architecture Search~\cite{NASZoph, liu2018darts} will likely lead to even more efficient models, which we leave to future work.

\section{Conclusion}

We demonstrate that for a constant computational budget, sparse convolutional networks are more accurate than dense ones; this corroborates the findings of~\cite{wavernn}, which demonstrated that for a set number of floating-point operations, sparse RNNs are more accurate than dense RNNs.
We enable the use of weight sparsity to accelerate state-of-the-art convolutional networks by providing fast SpMM kernels along with all necessary supporting kernels for ARM processors. On a Snapdragon 835 the sparse networks we present in this paper outperform their dense equivalents by $1.3-2.4\times$ in terms of wall clock time for a given top-1 accuracy while needing only $\approx66\%$ as many parameters -- equivalent to approximately one entire generation of improvement. By overturning the misconception that ``sparsity is slow'', we hope to open new avenues of research that would previously not be considered.

{\small
\bibliographystyle{ieee_fullname}
\bibliography{main}
}

\appendix
\section{Comparison with Intel MKL}

\begin{figure*}[!b]
\begin{center}
    \begin{subfloat}[][MBv1]{
    \centering
    \includegraphics[width=.45\linewidth]{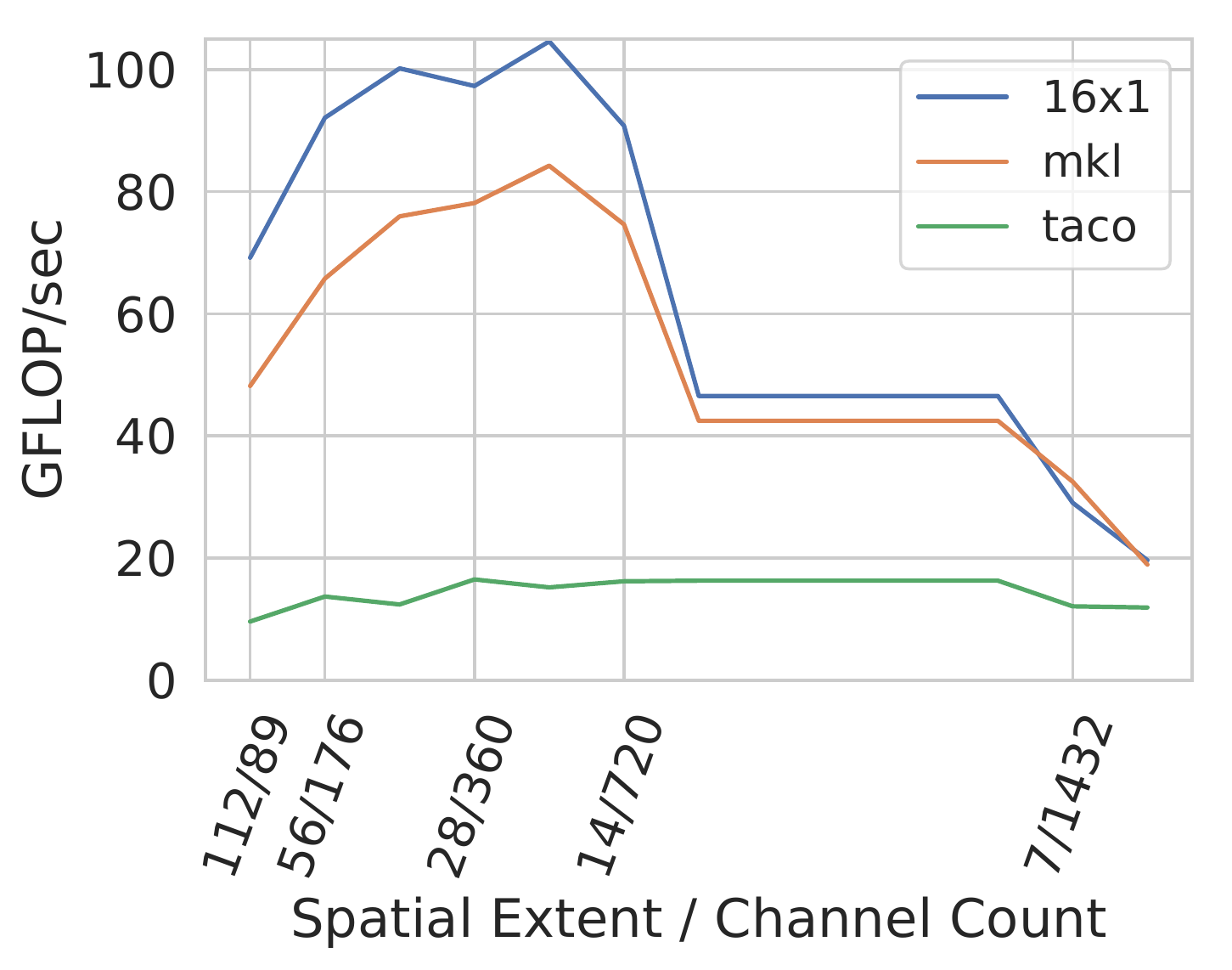}}
    \end{subfloat}
    ~
    \begin{subfloat}[][MBv2]{
    \centering
    \includegraphics[width=.45\linewidth]{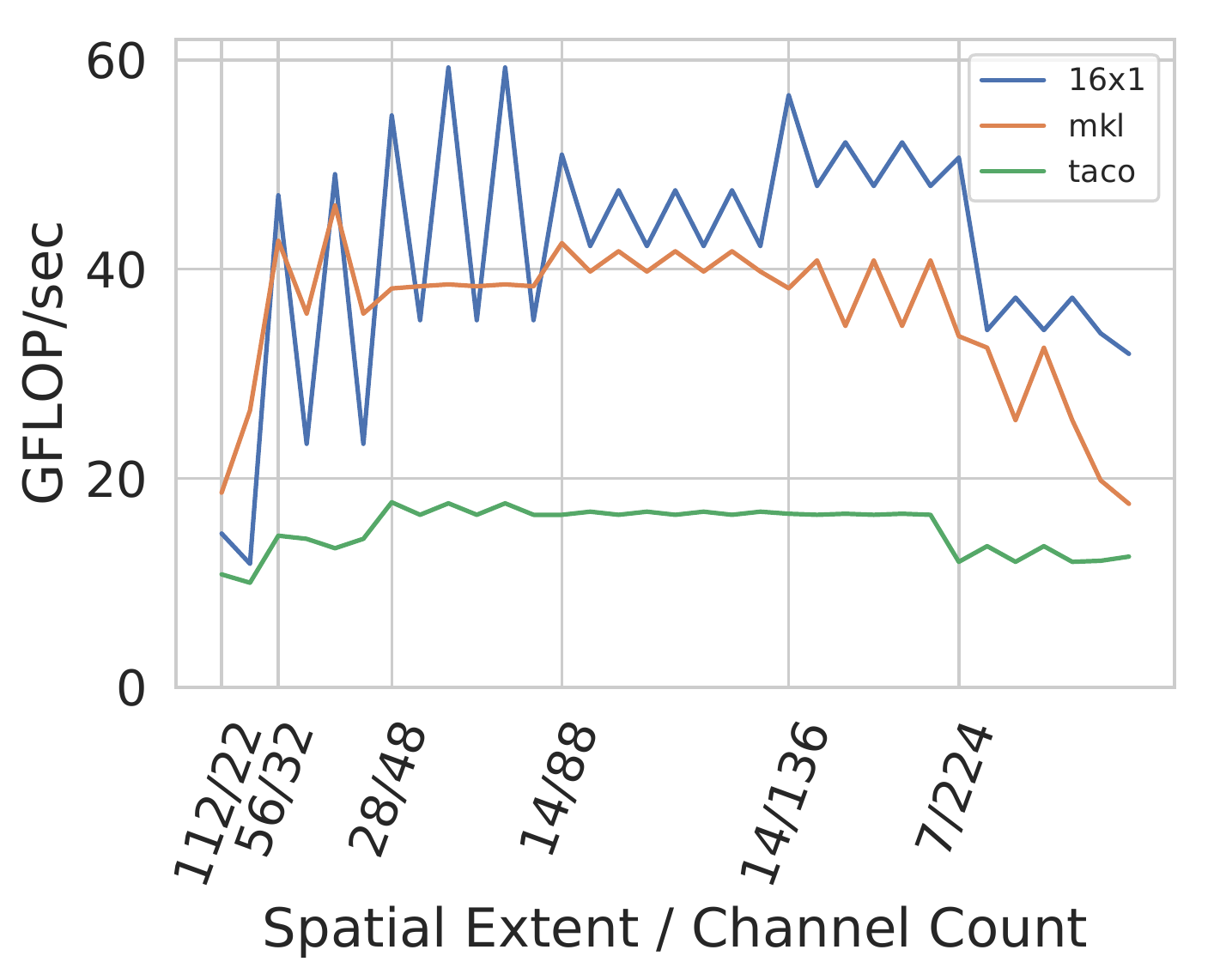}}
    \end{subfloat}
    \end{center}
    \caption{MBV1 (a) and MBv2 (b) achieved GFLOPs with increasing layer depth. Measurements taken on an Intel Xeon W-2135.}
    \label{fig:mbv1_mbv2_vsmkl}
\end{figure*}

We also implement our scheme with AVX512 intrinsics to compare with the Intel MKL.  With minimal tuning we find that it achieves a geometric mean speedup of 1.2$\times$ across all layers for both MBv1 and MBv2 compared with the MKL.  Results can be found in figure~\ref{fig:mbv1_mbv2_vsmkl}.

\begin{figure}
\begin{center}
    \includegraphics[width=\linewidth]{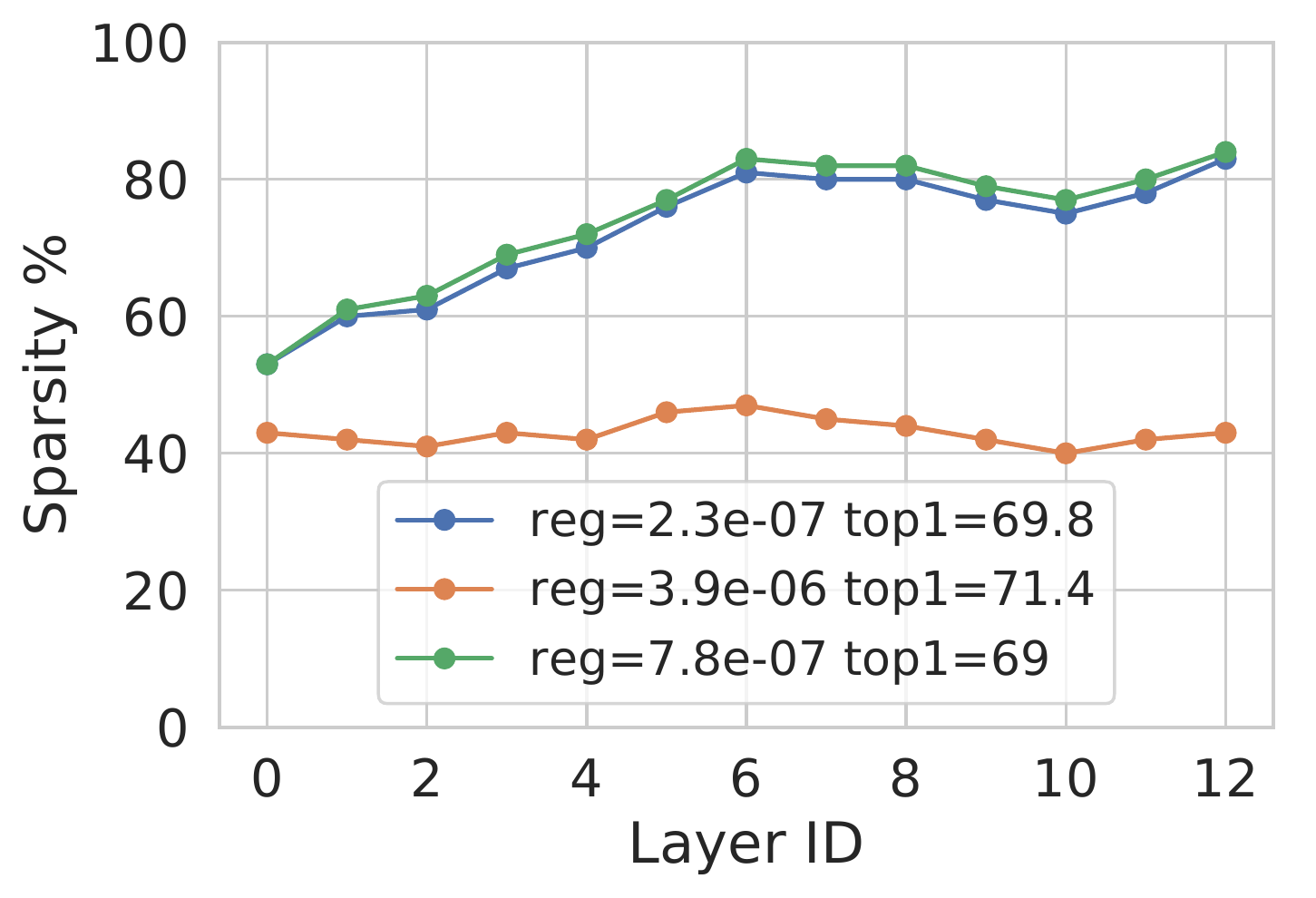}
\end{center}
\caption{MBv1 layer wise sparsities found with Variational Dropout. The curve with the highest regularization coefficient shows an interesting phenomena of collapse -- the model is actually less sparse and with more uniformity than models with lower regularization coefficients.  The layer just before the final spatial resolution decrease and channel doubling is preferred to be less than those before and after.  Early layers which have very few parameters are less sparse than later layers.}
\label{fig:mbv1_vd_layerwise_sparsity}
\end{figure}

\section{Non-Uniform Layerwise Sparsity with Variational Dropout}

Variational Dropout (see Molchanov et al. 2017) is a Bayesian technique for inducing weight (or activation) sparsity in neural networks.  It does not outperform pruning (see Gale et al. 2019) but it prunes globally, so it can induce non-uniform distributions of sparsity across layers without manual intervention. Parameter count and FLOP count do not have a 1:1 relationship in convolutional models -- removing parameters from layers when the spatial dimension is large reduces the FLOP count by significantly more than removing parameters from later layers when it is small. Combined with the fact that later layers tend to have many more parameters than early ones means that global pruning approaches such as VD which operate under a parameter constraint find models which have many more FLOPs than techniques such as pruning, which are generally used to induce near uniform sparsity in the model, even if both models have the same number of parameters.

An additional practical difficulty with using VD to prune models is that one cannot specify a desired final sparsity directly. Instead one must tune the weight of the KL penalty term so that after training and thresholding the resulting model has the desired sparsity. This makes finding a model with a desired sparsity a time consuming process.

Despite these limitations, we nonetheless find it insightful to examine the pattern of layer wise sparsity that is found by using VD to prune MBv1, MBv2 and EfficientNet, with the hope of using some of this knowledge when using magnitude pruning to find even more efficient models.

\begin{figure*}
\begin{center}
\begin{subfloat}[][MBv2]{
    \centering
    \includegraphics[width=.45\linewidth]{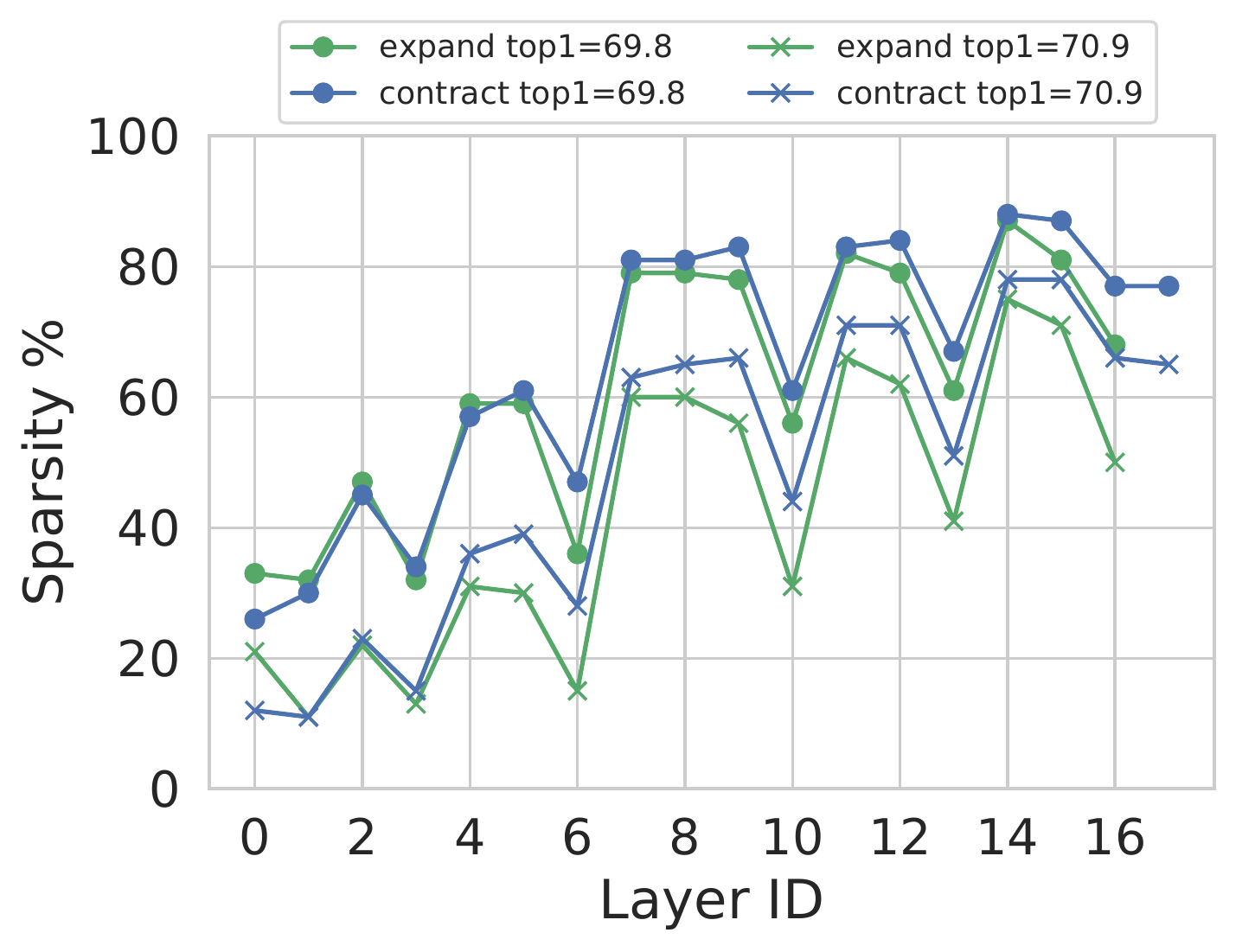}}
    \end{subfloat}
    ~
    \begin{subfloat}[][EfficientNet B0]{
    \centering
    \includegraphics[width=.45\linewidth]{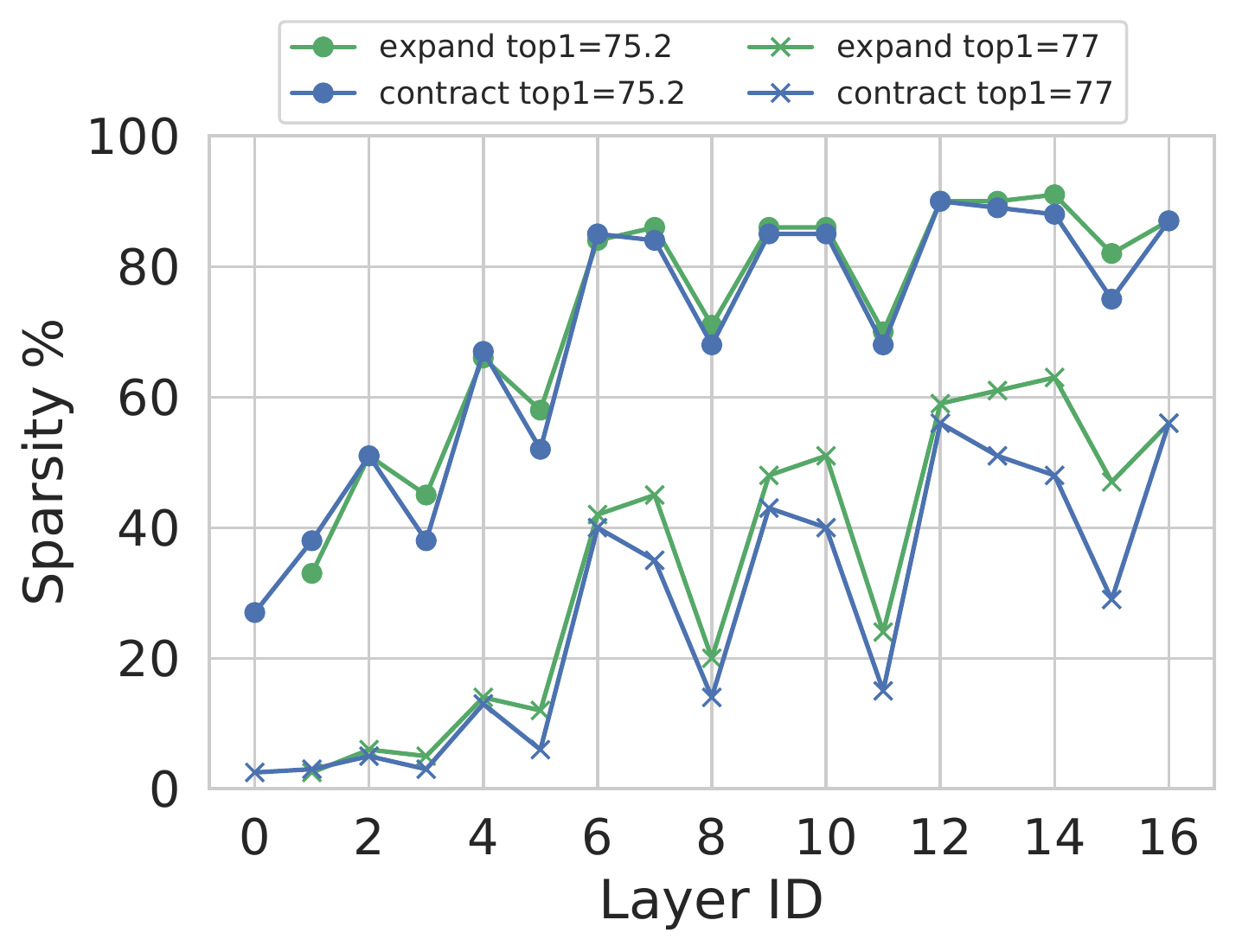}}
    \end{subfloat}
\end{center}
\caption{MBv2 layer wise sparsities found with Variational Dropout. Generally, it is preferred to keep the expansion matrices slightly less sparse than the contraction matrices.  There is a general trend to keeping to early layers with few parameters less sparse than later layers.  Within that general trend layers where the spatial resolution decreases and the channel count increases are much less sparse than would otherwise be expected.}
\label{fig:mbv2_en_vd_sparsity}
\end{figure*}

Results for MBv1 can be found in figure~\ref{fig:mbv1_vd_layerwise_sparsity} and results for MBv2 and EfficientNet in figure~\ref{fig:mbv2_en_vd_sparsity}.  As expected under a global parameter constraint, the layer wise sparsity increases with depth as the number of parameters per layer grows.  Interestingly, there is a trend, especially evident in MBv2 and EN models, that layers containing a spatial downsampling and concomitant channel increase should be less sparse than those before or after.  We were unable to take advantage of this information with hand tuning to find more efficient architectures than those with uniform sparsity, but it may provide a useful prior for a full architecture search.

\section{EfficientNet Plots}
\begin{figure*}
\begin{center}
\begin{subfloat}[][Block size effect on EN]{
    \centering
    \includegraphics[width=.45\linewidth]{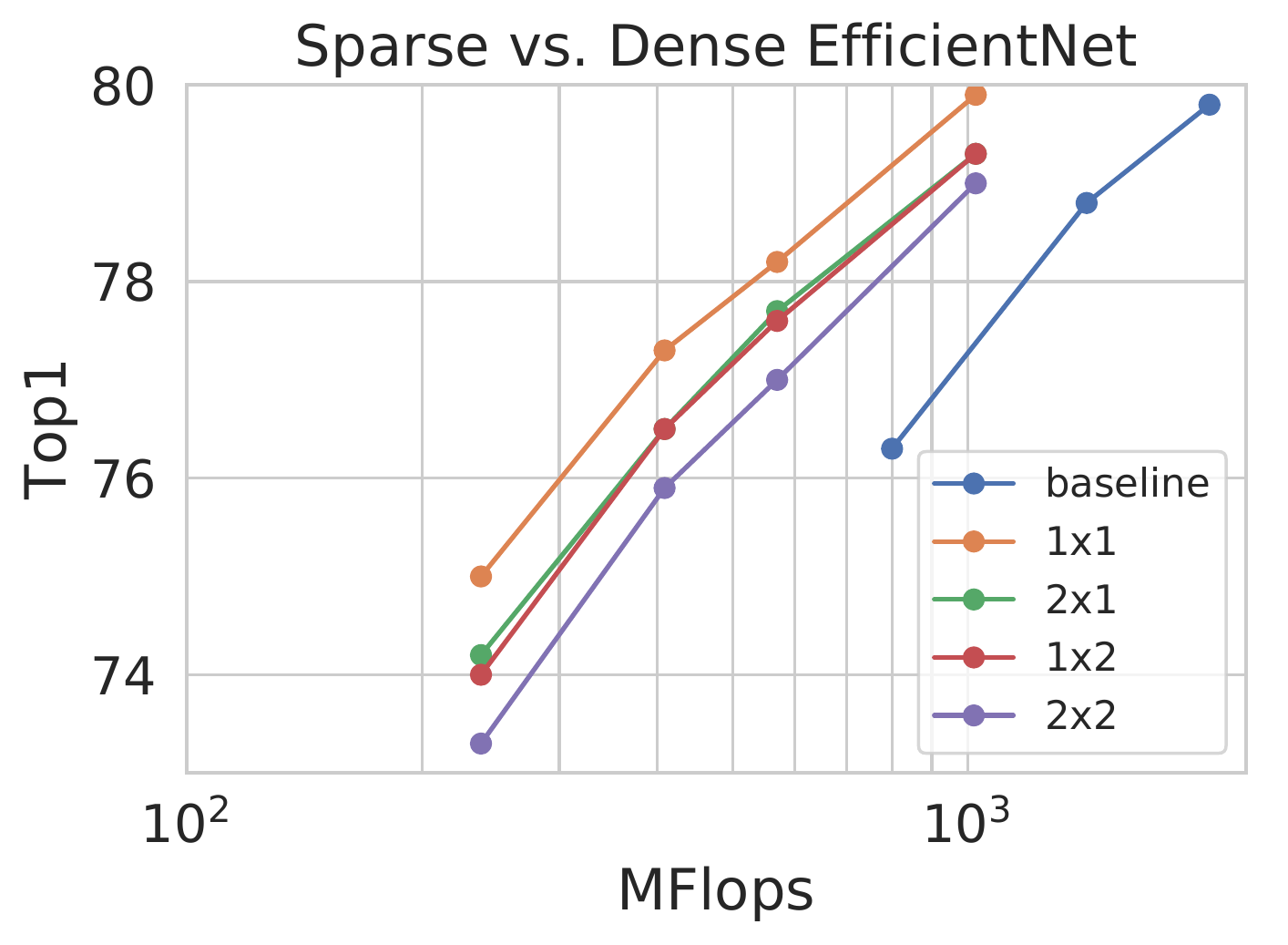}}
    \end{subfloat}
    ~
    \begin{subfloat}[][Sparsity level effect on EN]{
    \centering
    \includegraphics[width=.45\linewidth]{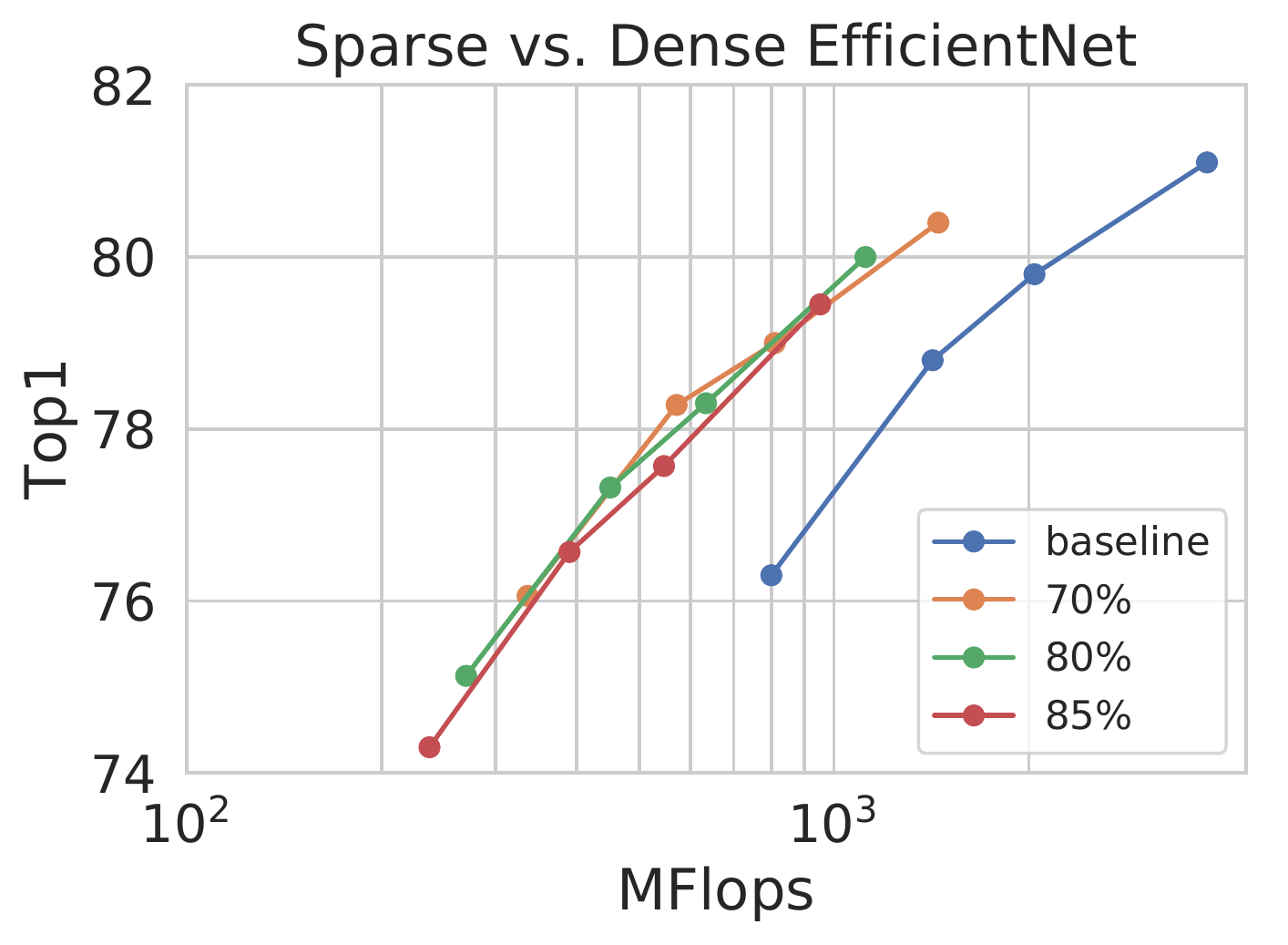}}
    \end{subfloat}
\end{center}
    \caption{EfficientNet behavior with: (a) different block sizes and (b) different sparsity levels.}
    \label{fig:en_plots}
\end{figure*}

Figures~\ref{fig:en_plots} contain the plots showing how EfficientNet behaves in the presence of blocks and for varying levels of sparsity.  It follows generally the same trend as the MobileNet models in the main text. The difference is that sparsities above 80\% do not seem to lead to more efficient models.

\end{document}